\documentclass[a4paper,twocolumn]{esapub2005} 
\usepackage{times}
\usepackage{url}

\usepackage{epsfig}
\usepackage{amssymb}
\usepackage{booktabs}
\usepackage{tabularx}
\usepackage{multirow}
\usepackage{subcaption}
\usepackage[export]{adjustbox}
\usepackage{float}
\usepackage{dashrule}
\usepackage{arydshln}
\usepackage[utf8x]{inputenc}
\usepackage[space]{grffile}
\usepackage{anyfontsize}
\usepackage{t1enc}
\usepackage{mathtools, nccmath}

\usepackage{amsmath,amsfonts}
\usepackage{algorithmic}
\usepackage{algorithm}
\usepackage{array}
\usepackage{textcomp}
\usepackage{stfloats}
\usepackage{url}
\usepackage{verbatim}
\usepackage{graphicx}
\usepackage{animate}
\usepackage{xcolor}

\newlength \mywidth
\newlength \myw
\title{Domain Adaptive Object Detection for Space Applications with Real-Time Constraints}
\author{Samet Hicsonmez}
\author{Abd El Rahman Shabayek}
\author{Arunkumar Rathinam}
\author{Djamila Aouada}
\affil{University of Luxembourg, Luxembourg, Luxembourg, (samet.hicsonmez,abdelrahman.shabayek,arunkumar.rathinam,djamila.aouada)@uni.lu}

\begin{document}


\maketitle

\begin{abstract}
Object detection is essential in space applications targeting Space Domain Awareness and also applications involving relative navigation scenarios.  Current deep learning models for Object Detection in space applications are often trained on synthetic data from simulators, however, the model performance drops significantly on real-world data due to the domain gap. However, domain adaptive object detection is an overlooked problem in the community. 
In this work, we first show the importance of domain adaptation and then explore Supervised Domain Adaptation (SDA) to reduce this gap using minimal labeled real data. We build on a recent semi-supervised adaptation method and tailor it for object detection. Our approach combines domain-invariant feature learning with a CNN-based domain discriminator and invariant risk minimization using a domain-independent regression head.
To meet real-time deployment needs, we test our method on a lightweight Single Shot Multibox Detector (SSD) with MobileNet backbone and on the more advanced Fully Convolutional One-Stage object detector (FCOS) with ResNet-50 backbone. We evaluated on two space datasets, SPEED+ and SPARK. The results show up to 20-point improvements in average precision (AP) with just 250 labeled real images. 
\end{abstract}    
\section{Introduction}
\label{sec:intro}

Vision-based object detection plays a central role in a wide range of space applications~\cite{dung2021spacecraft}, including Space Domain Awareness missions and also missions that involve relative navigation or autonomous rendezvous such as in-orbit servicing and active debris removal. Reliable detection of spacecraft or orbital objects is critical for enabling safe and autonomous operations in space. Modern object detection models, particularly those based on deep learning, require large amounts of annotated data to achieve high accuracy. However, collecting labeled data in the space domain is extremely difficult due to constraints such as limited accessibility, high operational costs, and the complexity of manual annotation under space conditions.

To mitigate the challenge of data scarcity, synthetic datasets~\cite{speed, spark} generated through simulation environments~\cite{olivares2023zero} have become the primary training source. Synthetic datasets are scalable and easy to generate and label. However, models trained solely on synthetic data often perform poorly when deployed in real-world scenarios~\cite{sda_1}. This performance drop is mainly attributed to the domain gap between the simulated training data (source) and the real orbital imagery (target), which includes differences in lighting, background, object textures, and noise characteristics.

Domain Adaptation (DA)~\cite{ben2006analysis} has emerged as a promising solution to bridge this gap. Unsupervised Domain Adaptation (UDA) methods~\cite{wilson2020survey} aim to align source and target domains without requiring labeled target images. Although they offer improvements over naive transfer, their ability to fully close the domain gap remains limited~\cite{aldi}. In contrast, semi-supervised domain adaptation methods~\cite{ssda_kr_4, zhou2023ssda}, which rely on a small set of labeled real images, have shown greater potential. Even limited real supervision, on the order of a few hundred annotated samples, can significantly enhance detection performance when properly integrated into the training process.

In semi-supervised domain adaptation (SDA), it is typically assumed that the target domain provides a small set of labeled samples together with a large collection of unlabeled data. While this assumption is reasonable in many application areas, it rarely holds in the space domain. Real-world space datasets usually contain only a few hundred images, making large-scale unlabeled target data unavailable and thus challenging the standard SDA setting.

Although object detection is a critical component in many space applications, the use of domain adaptation techniques for this task has received limited attention, until recently~\cite{ulmer2025important, zhang2025semi}. This may be due to the common assumption that the problem is relatively simple, as scenarios often involve a single target object against a relatively static or uniform background. However, both recent studies and our findings demonstrate that this assumption is considerably more challenging than commonly assumed.

In this paper, we build upon a recent semi-supervised domain adaptation approach and adapt it to become a Supervised Domain Adaptive (SDA) object detection. 
We employ a simple CNN-based domain discriminator to encourage feature alignment between domains and introduce a domain-independent regression head to enforce consistency in object localization. This setup allows the model to generalize better across domains, even with limited real data.

To evaluate the practicality of our approach for real-world deployment, we experiment with both lightweight and modern object detectors. Specifically, we use SSD~\cite{ssd} with a MobileNetv2~\cite{mobilenetv2} backbone for real-time performance and FCOS~\cite{fcos} with ResNet-50~\cite{resnet} for architectural generality. Experiments are conducted on two publicly available space datasets, SPEED+~\cite{speed} and SPARK~\cite{spark}. Our results show that the proposed approach significantly closes the domain gap and improves performance.

In Section~\ref{sec:rel_work}, we review existing domain adaptation methods in both real-world and space-related contexts. Section~\ref{sec:method} provides a detailed description of our proposed approach. Experimental results are presented in Section~\ref{sec:exp}, followed by concluding remarks in Section~\ref{sec:conc}.

\section{Related Work}
\label{sec:rel_work}

In this section, we first explain the main approaches for domain adaptation in natural images and then describe the methods targeted for space applications in the context of \textit{object detection}. 

\subsection{Domain Adaptation for Natural Images}

Unsupervised approaches~\cite{kennerley2024cat, chen2018domain, cai2019exploring, chen2022learning, cao2023contrastive, deng2021unbiased, krishna2023mila} currently dominate research in Domain Adaptive Object Detection (DAOD). Only a few methods explore semi-supervised DA~\cite{ssda_kr_4, ijcai2022p213}, and there is virtually no work focused on SDA training. However, unsupervised or semi-supervised methods can often be adapted to supervised settings, potentially yielding better performance in the target domain. Therefore, the following review focuses on UDA methods.

Following the taxonomy presented in a recent survey on UDA for object detection~\cite{oza2023unsupervised}, the dominant DAOD approaches can be broadly grouped into three categories: 1) adversarial feature learning~\cite{chen2018domain, saito2019strong, sindagi2020prior, hsu2020every, Zheng_2020_CVPR, vs2021mega, zhao2022task}, 2) image-to-image translation~\cite{progressive, zhang2019cycle, DBLP:conf/bmvc/RodriguezM19, harmonizing, cdtd, LIN2023109416, Yu_2022_WACV}, and 3) mean teacher training~\cite{cai2019exploring, deng2021unbiased, li2022cross, kennerley20232pcnet, kennerley2024cat}. 

\textbf{Adversarial feature learning} aims to match the features coming from the source and target domains with the help of a discriminator. In this setting, usually a single object detector model (e.g. Faster-RCNN~\cite{ren2015faster}, SSD~\cite{ssd} or FCOS~\cite{fcos}) is trained using images from both domains, and depending on the method, one or many domain discriminators are employed with GRLs.

The main advantage of this approach is that it improves the baseline source-only training with a very small number of parameters, i.e. shallow domain classifiers. However, since the single detection network is trained using images from both the source and target domains, the network cannot capture specific features of the target, which limits overall performance.

\textbf{Image to image translation} aims to close the gap between the source and target domains by using an unpaired image-to-image translation (im2im) model such as CycleGAN~\cite{cyclegan}, UNIT~\cite{unit}, MUNIT~\cite{munit} or CUT~\cite{park2020contrastive}. Generative Adversarial Networks (GAN)~\cite{gan}-based im2im models learn a mapping from a source domain $X$ to a target domain $Y$. Paired im2im models have ground truth pairs in both domains such as an image and its edge map, and unpaired im2im models have no ground truth pairs and employ unsupervised training.
After translation, an intermediate image representation that resembles the target domain more than the initial source domain is obtained, and since the method works at the pixel level, it is possible to use the source annotations on the intermediate dataset. 

The image translation part of this architecture is trained offline, which adds little or no overhead to the main object detection method training, and the target translated images show better resemblance than using special image augmentations, for example, contrast adjustment for day to night domain adaptation. In addition, im2im is widely used in the training of other DAOD approaches.

\textbf{Mean teacher} (MT) based training is proposed to overcome the limitations of training a single method (i.e. biasing towards the source dataset) using both source and target datasets. In this setting, a student model is trained using labeled source data with supervised training, the teacher model is trained unsupervised using only the target dataset, and the student model updates the weights of the teacher model by exponential moving average. Finally, inconsistency between the detections of the student and teacher models is penalized to support the learning. The weights of the teacher model are used for inference on the target dataset. 

MT based methods show superior performance compared to the above two approaches. Most, if not all, of the MT methods employ an offline trained im2im method to augment the source and target datasets. Moreover, the MT's teacher network is trained using only target images, making the training more stable. Adversarial learning is also included in some recent MT approaches~\cite{li2022cross, kennerley2024cat}.

\subsection{Domain Adaptation for Space Images}

Domain adaptation (DA) for space imagery has gained attention because of the difficulty of collecting real space data. However, existing DA research has focused primarily on pose estimation tasks~\cite{sda_1, park2023robust, liu2024revisiting, perez2023spacecraft, wang2023bridging}, while the object detection task remains largely overlooked. Recently, fANOVA~\cite{ulmer2025important} explored the impact of data augmentations in the UDA setting. Their framework identifies optimal combinations of image augmentations to significantly enhance source-only training performance. Another recent work~\cite{zhang2025semi} incorporates DA using image-to-image translation. The authors train a translation model to generate target-style images, which are then used to improve the generalization of the model across domains. 

In contrast to these works, we propose a DA framework in a more realistic SDA setting, where a limited number of labeled target-domain images are available. Our approach explicitly leverages this limited supervision to improve object detection performance under domain shift.

\section{Method}
\label{sec:method}

The proposed approach is built on top of a recent semi-supervised domain adaptation framework titled LIRR~\cite{ssda_kr_4} (Learning Invariant Representations and Risks). LIRR proposes to jointly learn invariant representations and risks at the same time to better mitigate the accuracy discrepancy across domains. LIRR presents a generic framework which could be used in both classification and regression tasks. In our work, we utilize it for a regression task, i.e. object detection, in a fully-supervised setting considering all the target data is labeled. 

\textbf{Invariant representation} ensures that the features of the images of the source and target domain extracted using a feature extractor backbone $g(.)$ are not separable and is formulated as follows:

\begin{align}
\mathcal{L}_{\text{rep}}(g, \mathcal{C}) =\  
& \mathbb{E}_{X \sim \mathcal{D}_S(X)} [\log(\mathcal{C}(g(X)))] \notag \\
+\, & \mathbb{E}_{X \sim \mathcal{D}_T(X)} [\log(1 - \mathcal{C}(g(X)))].
\end{align}

where $C$ is the domain classifier. 

\textbf{Invariant risk} aims to minimize domain-invariant task losses, i.e. cross-entropy losses of domain-invariant
predictor $f_i$ and domain-dependent predictor $f_d$. Considering that we have a domain-invariant loss $L_i$ formulated as follows:
\begin{equation}
\min_{g, f_i} \mathcal{L}_i = \mathbb{E}_{(x, y) \sim \mathcal{D}_S, \mathcal{D}_{\tilde{T}}} \left[ L\left(y, f_i(g(x))\right) \right],
\end{equation}

where ($x$, $y$) pairs denote input data and its ground truth label, and $L$ denotes task-dependent loss. In our case, $L$ is the sum of the classification (cross entropy) and localization (smooth $L_1$) losses.
The domain-dependent loss $L_d$ is formulated as follows:
\begin{equation}
\min_{g, f_d} \mathcal{L}_d = \mathbb{E}_{d \sim \mathcal{D} \,\, (x, y) \sim \mathcal{D}_d} \left[ L\left(y, f_d(g(x), d)\right) \right],
\end{equation}
where the additional $d$ signifies the domain from which the data is sampled.

Thus, the invariant risk loss is formulated as follows:

\begin{equation}
\min_{g, f_i} \max_{f_d} \mathcal{L}_{\text{risk}} = \mathcal{L}_i + \lambda_{\text{risk}} (\mathcal{L}_i - \mathcal{L}_d).
\end{equation}

\textbf{The final loss function} used to train the methods could be described as:

\begin{align}
\min_{g, f_i} \max_{\mathcal{C}, f_d} \mathcal{L}_{\text{LIRR}}(g, f_i, f_d, \mathcal{C})
= \;& \mathcal{L}_{\text{risk}}(g, f_i, f_d) \notag \\
& + \lambda_{\text{rep}} \mathcal{L}_{\text{rep}}(g, \mathcal{C}).
\end{align}

where $\lambda_{\text{rep}}$ is set to $0.1$, and $\lambda_{\text{risk}}$ is set to $1.0$ in our experiments.

Note that all the equations are directly taken from LIRR~\cite{ssda_kr_4} and presented here for completeness. 

\section{Experiments}
\label{sec:exp}

\begin{table}
    \centering
    \caption{Details of SPARK and SPEED+ datasets used in the experiments. }
    \setlength\tabcolsep{4.0pt}
    \resizebox{1.0\columnwidth}{!}{
        \begin{tabular}{cccccc}
        \toprule
\multirow{2}{*}{Dataset}  & \multicolumn{2}{c}{Source} & & \multicolumn{2}{c}{Target}\\
        \cline{2-3} \cline{5-6} 
        & {Train} & {Val} &  & {Train} & {Val} \\
        \toprule
SPARK~\cite{spark}   & 22500 &  7500 & & 250/500 & 1600\\ 
SPEED+~\cite{speed} Sunlamp & 47966 &	11994 & & 250/500 & 2200  \\
SPEED+~\cite{speed} Lightbox & 47966 &	11994 & & 250/500 & 6200  \\
        \bottomrule
        \end{tabular}}
    \label{tab:dataset_stats}
\end{table}

\begin{table*}[t]
\begin{center}
\begin{tabular}{llccccc|cccc}
\toprule 
& OD & {Method} & Ims & $AP$ & $AP_{50}$ &  $AP_{75}$ & Ims & $AP$ & $AP_{50}$ &  $AP_{75}$ \\
\midrule 

\parbox[t]{2mm}{\multirow{6}{*}{\rotatebox[origin=c]{90}{Sunlamp}}} & 
\parbox[t]{2mm}{\multirow{3}{*}{\rotatebox[origin=c]{90}{SSD}}} & 
Source Only & -  & 44.4 &  85.2 & 42.1  & -  & 44.4 &  85.2 & 42.1 \\
& & Oracle & 250 & 53.0	& 98.0	& 50.8 & 500  & 63.4	& 98.9	& 76.5 \\
& & SDA & 250 & 79.4	& 99.0	& 97.6 & 500  & 82.0	& 99.0	& 97.7  \\

\cmidrule(lr){2-11} 
& \parbox[t]{2mm}{\multirow{3}{*}{\rotatebox[origin=c]{90}{FCOS}}} & 
 Source Only & -  & 52.1 &  95.1 &  52.6  & -  & 52.1 &  95.1 &  52.6 \\
& & Oracle & 250 & 72.8 & 98.9	& 90.8 & 500  & 77.1	& 99.0	& 94.8 \\
& & SDA & 250 & 52.1 &  98.9 &  81.7  & 500  & 66.3 &  98.9 &  81.7 \\

\midrule 

\parbox[t]{2mm}{\multirow{6}{*}{\rotatebox[origin=c]{90}{Lightbox}}} & 
\parbox[t]{2mm}{\multirow{3}{*}{\rotatebox[origin=c]{90}{SSD}}} & 
Source Only & -  & 26.2 &  81.1 &  35.8  & -  & 26.2 &  81.1 &  35.8 \\
& & Oracle & 250 & 46.8	& 95.8	& 37.3 & 500  & 60.5	& 98.6	& 69.6 \\
& & SDA& 250 & 83.4	& 99.0	& 96.3 & 500  & 85.7	& 99.0	& 97.6  \\

\cmidrule(lr){2-11} 
& \parbox[t]{2mm}{\multirow{3}{*}{\rotatebox[origin=c]{90}{FCOS}}} & 
Source Only & -  & 11.2 &  36.6 &  4.2  & -  & 11.2 &  36.6 &  4.2  \\
& & Oracle & 250 & 56.8	& 94.8	& 62.0 & 500  & 65.8	& 97.2 & 76.0 \\
& & SDA  & 250 & 54.6 &  94.6 &  60.7  & 500  & 61.4 &  96.3 &  70.1 \\

\bottomrule 
\end{tabular}
\end{center}
\caption{Results on the SPEED+ dataset for the sunlamp and lightbox subsets.}
\label{tab:speed_scores}
\end{table*}

\subsection{Datasets}
SPEED+~\cite{speed} is a well-known dataset in the spacecraft pose estimation domain that features synthetic images generated from a simulator along with two distinct hardware-in-the-loop (HIL) image domains, \textit{sunlamp} and \textit{lightbox}. The \textit{lightbox} domain contains 6,740 images of the mockup of the Tango spacecraft, illuminated using albedo light boxes to simulate diffuse lighting conditions typical of Earth orbit. We split this dataset into train and test sets. The test set contains $6200$ images, and the remaining images were sampled into two train subsets with $250$ and $500$ images from the remaining images. The \textit{sunlamp} domain includes 2,791 images of the same mockup, captured under a metal halide arc lamp to replicate direct sunlight. These domain-specific lighting conditions make SPEED+ a valuable benchmark for studying domain adaptation in space-based vision tasks. Similarly, we split the images in the sunlamp domain into a test set with $2200$ images and two training sets with $250$ and $500$ images. Note that in both domains, the larger train set completely overlaps with the smaller training split.
Additionally, this dataset contains close to $60,000$ synthetic images.

SPARK~\cite{spark} dataset is collected in the SnT Zero-G Lab~\cite{olivares2023zero} at the Interdisciplinary Centre for Security, Reliability, and Trust (SnT), University of Luxembourg. The source domain comprises $100$ trajectory groups, each containing $300$ labeled synthetic images. The target domain includes four real trajectory sequences, RT001, RT002, RT003, and RT004, consisting of 681, 424, 678, and 340 time-sequential images, respectively. We created non-overlapping train and test splits for SPARK as well. 

The statistics of the dataset in terms of the number of images in each split are presented in Table~\ref{tab:dataset_stats}.

\subsection{Setup}
In order to evaluate the performance of the proposed method, we devise a three-stage evaluation pipeline, see Table~\ref{tab:spark_scores} and~\ref{tab:speed_scores}. First, we quantify the domain gap by training a model using only the source (synthetic) training dataset and evaluate it on the target (lab) test set. This experiment is labeled as \textit{source only}. Second, we train a model using only the target training set to measure performance without domain adaptation, and it is labeled as \textit{Oracle}. Finally, we train our proposed model using both the source and target training sets to see the effect of domain adaptation. Note that all models are evaluated on the newly created target test set mentioned in the above section.    

To evaluate our approach under limited supervision, we select subsets of $250$ and $500$ labeled real images from both datasets to train the Oracle and SDA models. We conduct experiments using two object detection frameworks: SSD~\cite{ssd} with a MobileNetV2~\cite{mobilenetv2} backbone, chosen for its suitability in resource-constrained deployment scenarios, and FCOS~\cite{fcos} with a ResNet-50~\cite{resnet} backbone, used for general purpose performance evaluation. We adapt our approach to the official implementation of each method. 

The images are resized to $300\times300$ pixels for SSD training, and for the FCOS, the images are resized to have a shorter side of $800$ pixels. 
During inference, the images are resized similarly to the training, and we do not employ any data augmentation.

\subsection{Metrics}
We evaluated object detection performance using standard COCO-style~\cite{lin2014microsoft} metrics: Average Precision (AP), $AP_{50}$, and $AP_{75}$. The AP score measures the mean average precision over multiple IoU thresholds (ranging from $0.50$ to $0.95$ in $0.05$ increments), providing a comprehensive assessment of overall detection quality. $AP_{50}$ reports average precision at a fixed IoU threshold of $0.50$ and reflects the model's ability to locate objects coarsely. In contrast, $AP_{75}$ evaluates the performance at a stricter IoU threshold of $0.75$, highlighting the precise localization capability. These metrics allow us to assess both robustness and localization accuracy under domain-shift conditions.

\begin{table*}
\begin{center}
\begin{tabular}{llccccc|cccc}
\toprule 
& OD &  {Method} & Ims & $AP$ & $AP_{50}$ &  $AP_{75}$ & Ims & $AP$ & $AP_{50}$ &  $AP_{75}$ \\
\midrule 

\parbox[t]{2mm}{\multirow{6}{*}{\rotatebox[origin=c]{90}{Spark}}} & 
\parbox[t]{2mm}{\multirow{3}{*}{\rotatebox[origin=c]{90}{SSD}}} & 
Source Only & -  & 27.9 &  88.1 &  7.1  & -  & 27.9 &  88.1 &  7.1 \\
& & Oracle & 250 & 80.9	& 98.9	& 98.9 & 500  & 86.2 & 98.9	& 98.9 \\
& & SDA & 250 & 86.6 & 98.9 & 98.9 & 500 & 87.2 & 98.9	& 98.9 \\

\cmidrule(lr){2-11} 

& \parbox[t]{2mm}{\multirow{3}{*}{\rotatebox[origin=c]{90}{FCOS}}} & 
Source Only & -  & 27.2 &  85.6 &  6.5  & -  & 27.2 &  85.6 &  6.5 \\
& & Oracle & 250 & 95.9 & 99.0 & 99.0 & 500  & 97.9 & 99.0 & 99.0 \\
& & SDA & 250 & 96.6 &  99.0 &  99.0  & 500  & 98.0 & 99.0 &  99.0 \\

\bottomrule 
\end{tabular}
\end{center}
\caption{Results on the SPARK dataset.}
\label{tab:spark_scores}
\end{table*}

\subsection{Quantitative Results}

We present SPEED+~\cite{speed} and SPARK~\cite{spark} quantitative results in Table~\ref{tab:speed_scores} and Table~\ref{tab:spark_scores}, respectively. In each table, we show the metrics for the $250$ and $500$ images, and for SSD~\cite{ssd} and FCOS~\cite{fcos} object detectors.    

First, the \textit{source only} baseline shows the severity of the domain gap in both datasets, as the performance using only the labeled source dataset achieves dramatically low performance on the target data. 

On the SPEED+ dataset (Table~\ref{tab:speed_scores}), applying our SDA strategy to the SSD detector yields a substantial gain, exceeding 20 points in AP, over the \textit{Oracle} baseline. This highlights the critical role of domain adaptation, particularly for shallow detectors. In contrast, deeper architectures such as FCOS exhibit degraded performance when trained with source-domain data, with their \textit{Oracle} results surpassing SDA in both the Sunlamp and Lightbox splits.

A further noteworthy finding is that SSD with SDA outperforms the \textit{Oracle} performance of FCOS on both splits by more than 5 AP points. This underscores the importance of domain adaptation when using lightweight detectors, likely because SDA mitigates overfitting in models with fewer parameters.

On the SPARK dataset (Table~\ref{tab:spark_scores}), we observe performance gains for both detectors under low-data regimes. However, as the number of training images increases, the gap between the \textit{Oracle} and SDA models diminishes, and their performances converge to a comparable level.

\renewcommand{\arraystretch}{1}
\begin{figure*}[b]
\captionsetup[subfigure]{labelformat=empty}
\centering
\begin{tabular}{lcc|cc}

{\textit{\textbf{{}}}} & {\textit{\textbf{{ORACLE}}}} & {\textit{\textbf{{SDA}}}} & {\textit{\textbf{{ORACLE}}}} & {\textit{\textbf{{SDA}}}}  \\ 

{\rotatebox[origin=t]{90}{\textit{\textbf{Lightbox}}}}  &
\includegraphics[width=\myw,  ,valign=m, keepaspectratio,] {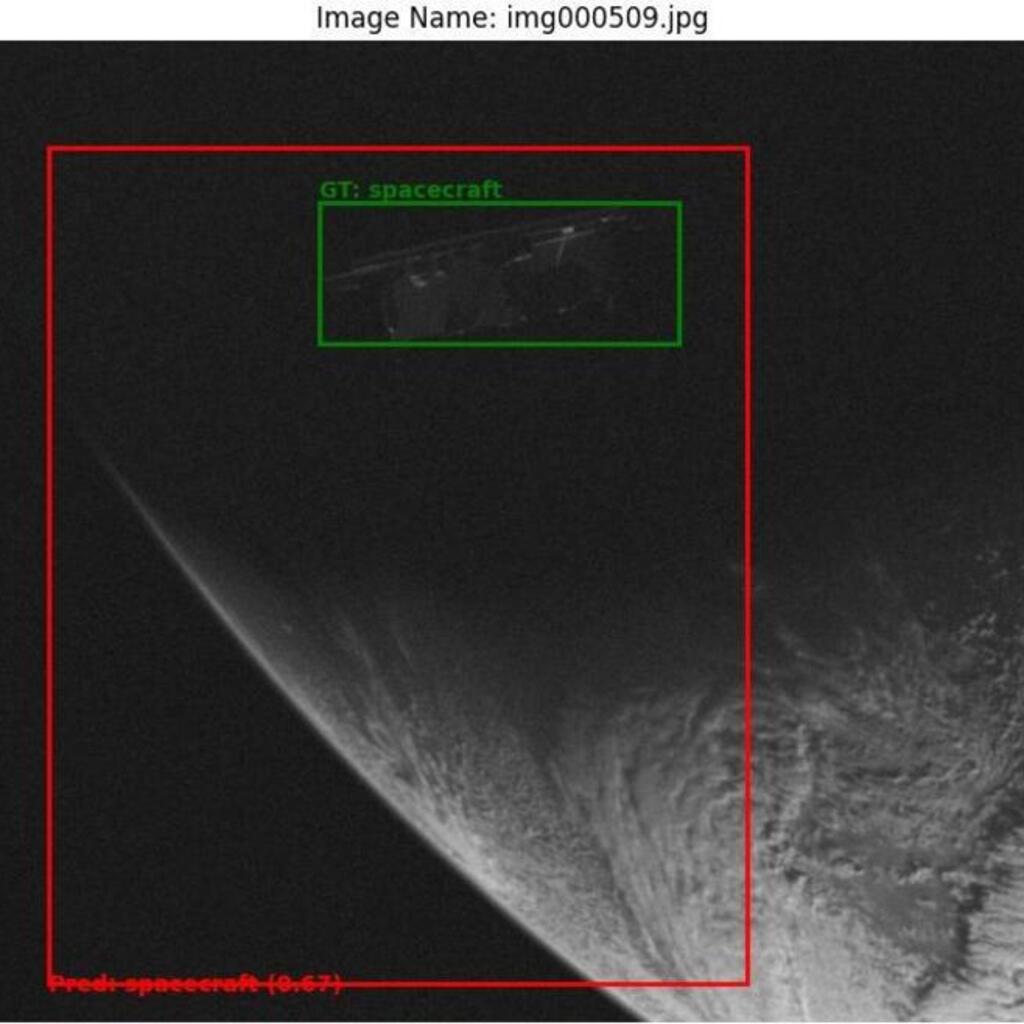} &
\includegraphics[width=\myw,  ,valign=m, keepaspectratio,] {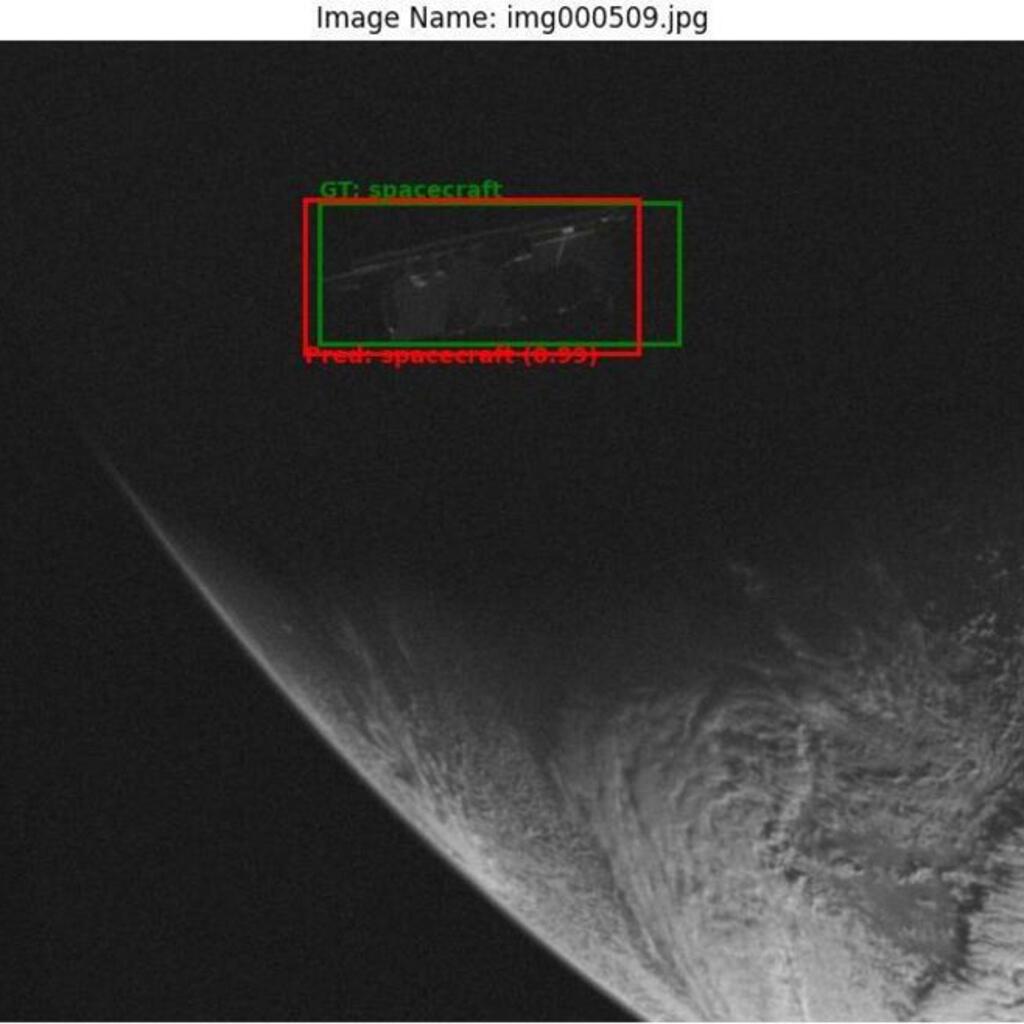} &

\includegraphics[width=\myw,  ,valign=m, keepaspectratio,] {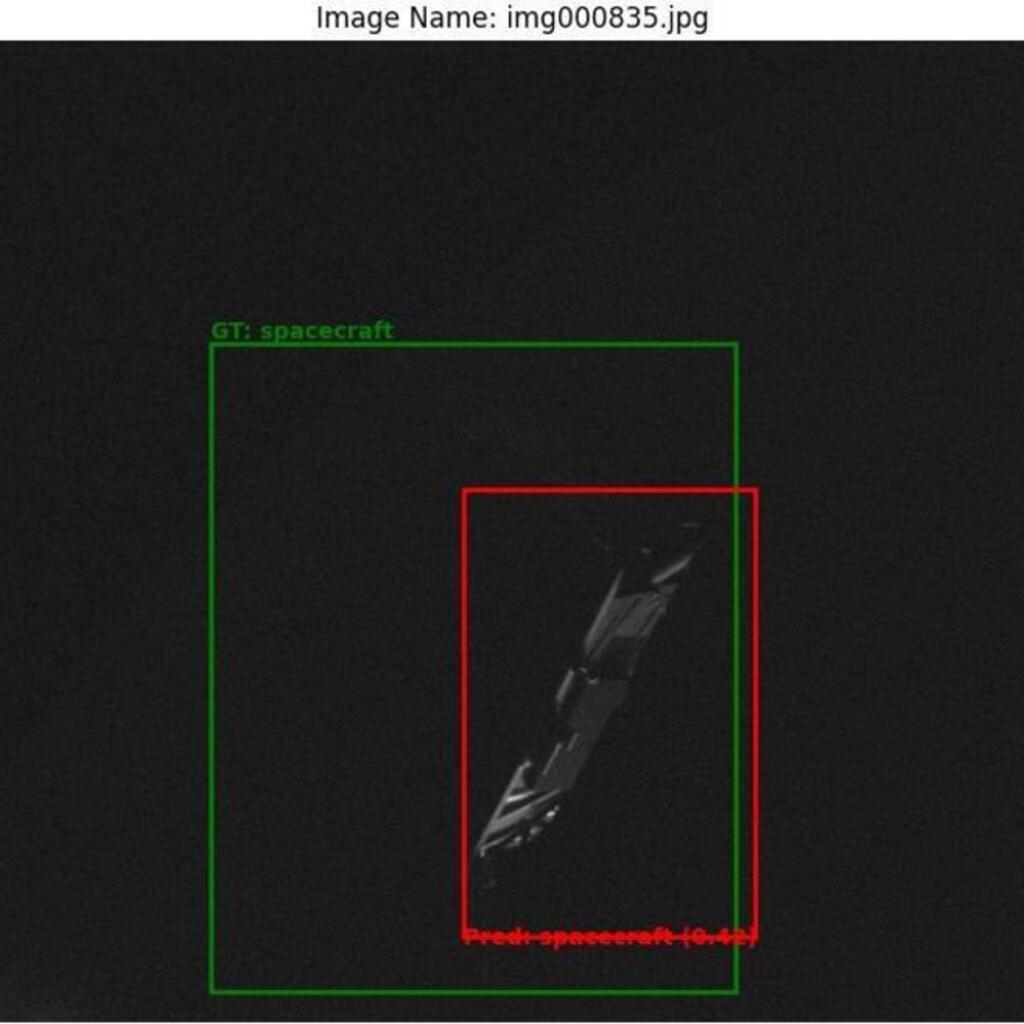} & 
\includegraphics[width=\myw,  ,valign=m, keepaspectratio,] {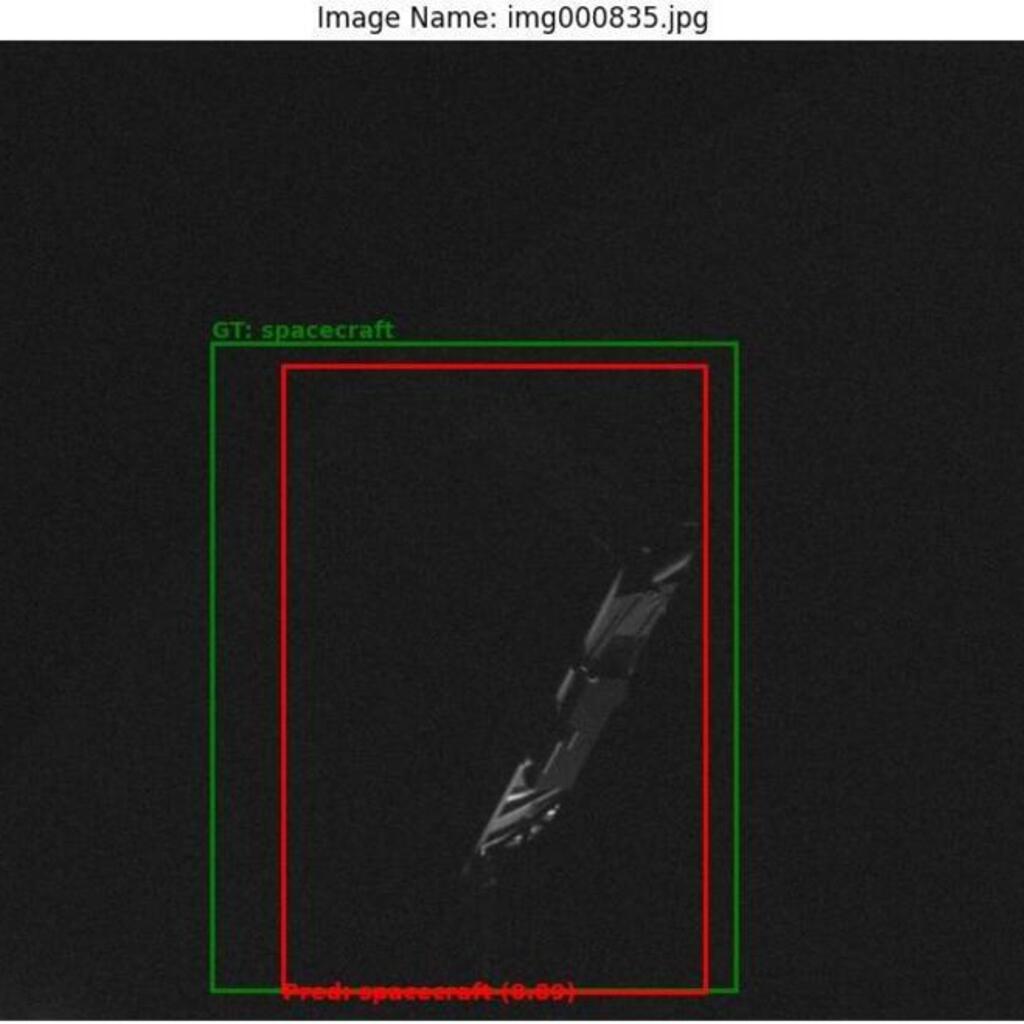}\\

{\rotatebox[origin=t]{90}{\textit{\textbf{Lightbox}}}}  &
\includegraphics[width=\myw,  ,valign=m, keepaspectratio,] {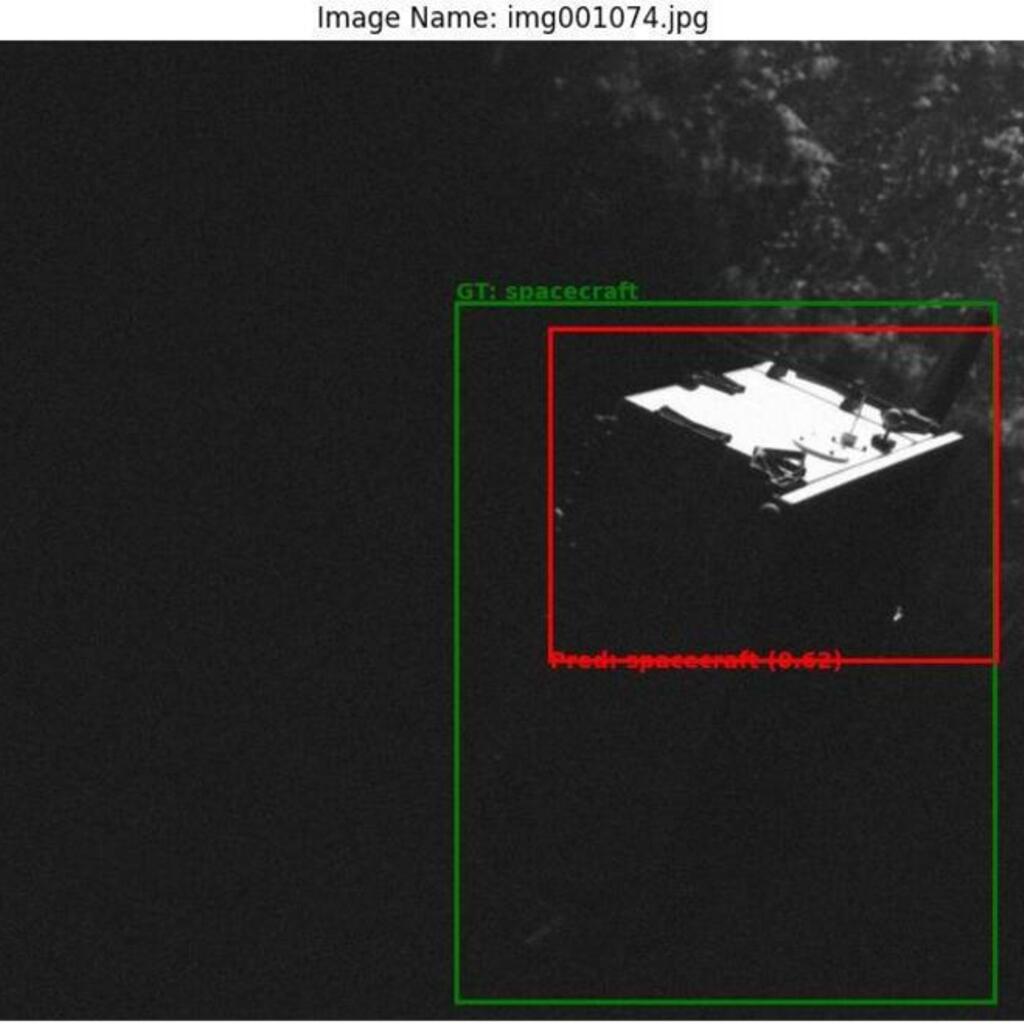} & 
\includegraphics[width=\myw,  ,valign=m, keepaspectratio,] {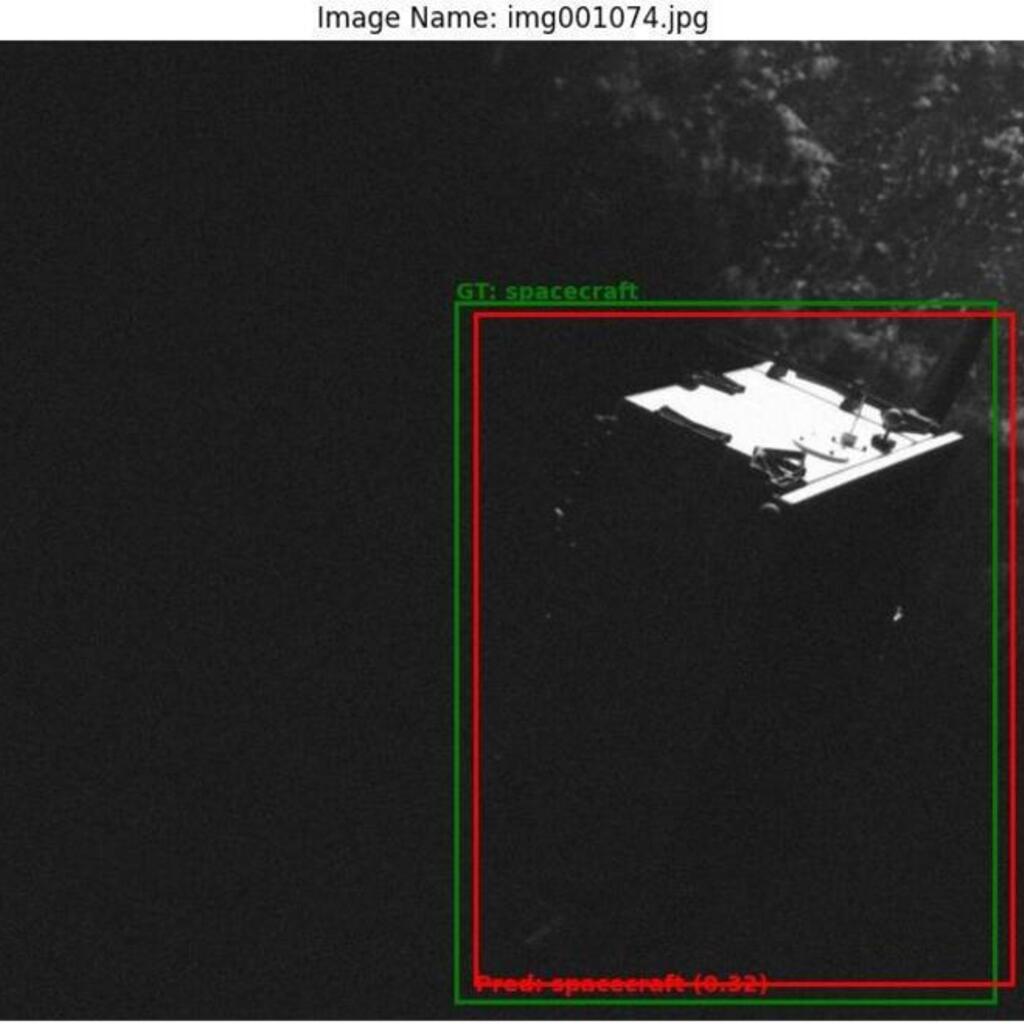} & 

\includegraphics[width=\myw,  ,valign=m, keepaspectratio,] {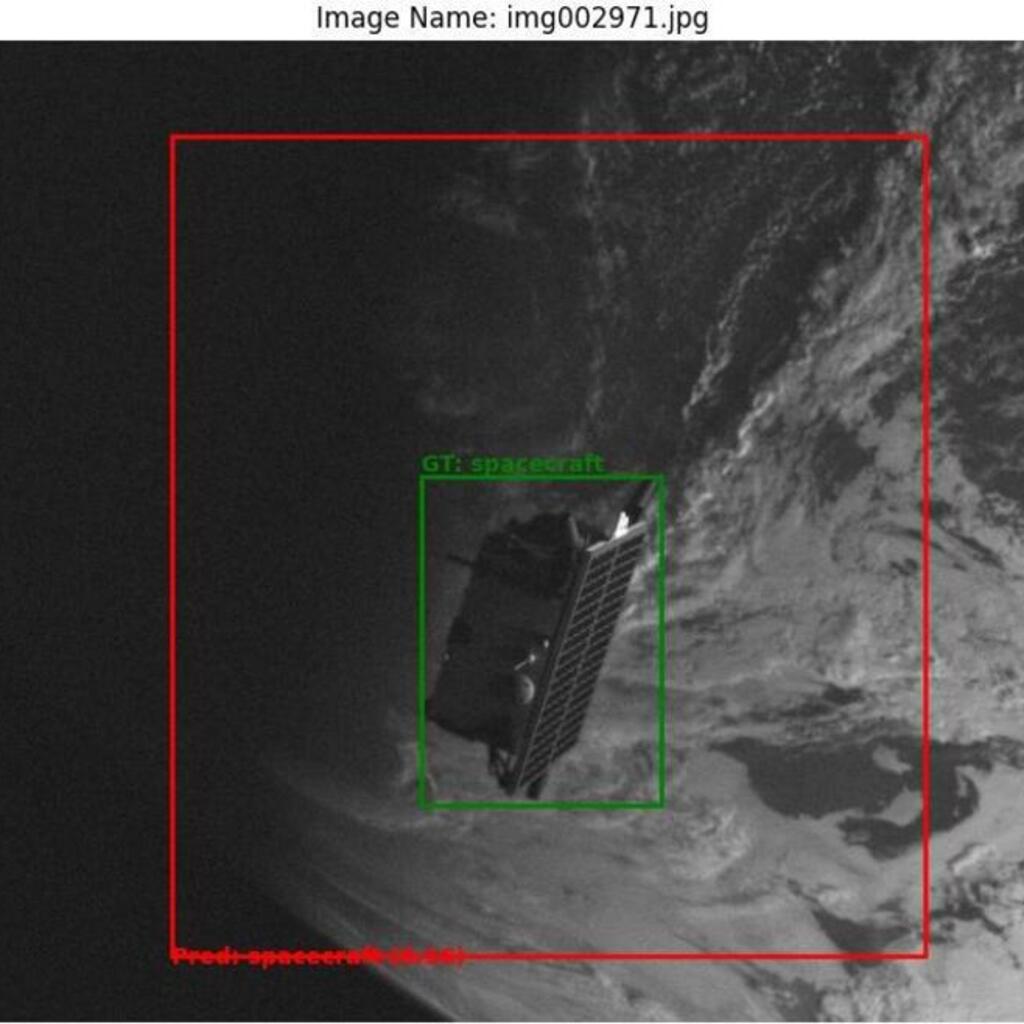} & 
\includegraphics[width=\myw,  ,valign=m, keepaspectratio,] {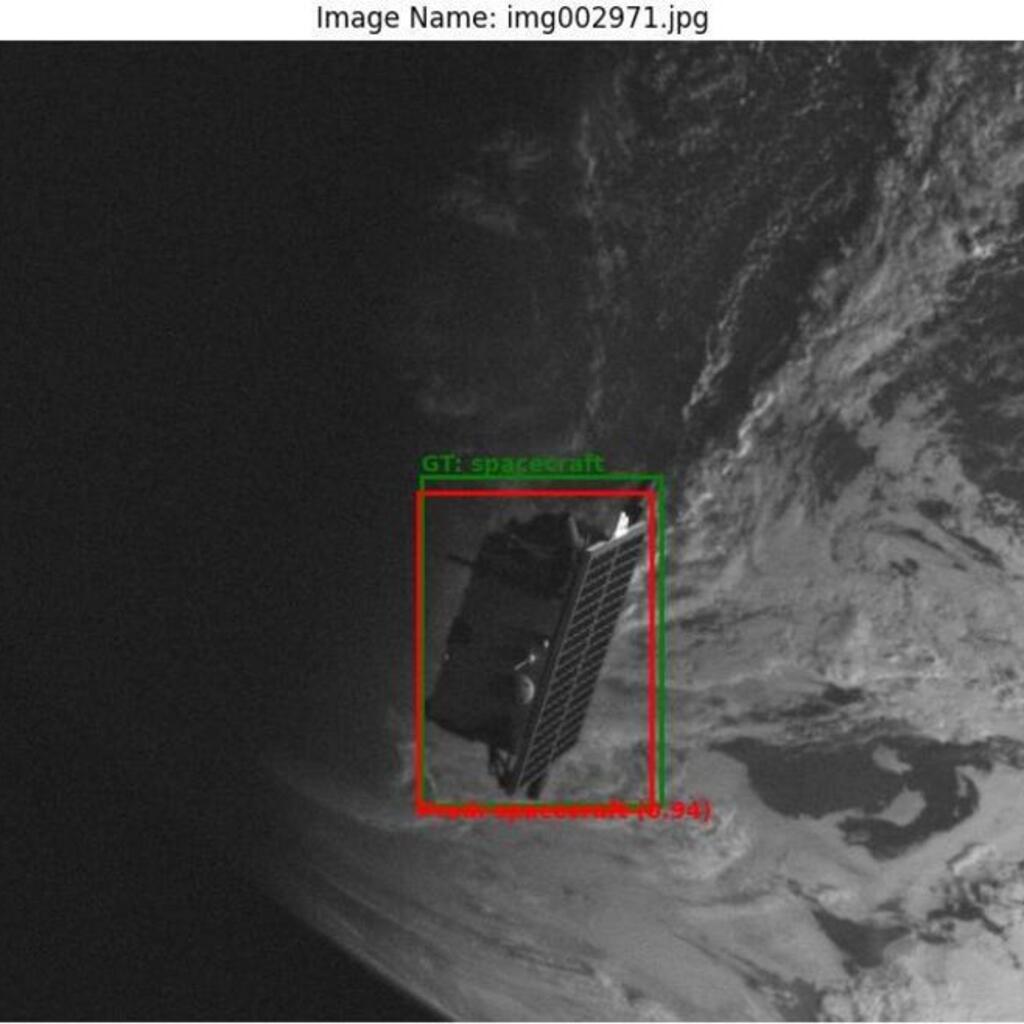}\\

{\rotatebox[origin=t]{90}{\textit{\textbf{Sunlamp}}}}  &
\includegraphics[width=\myw,  ,valign=m, keepaspectratio,] {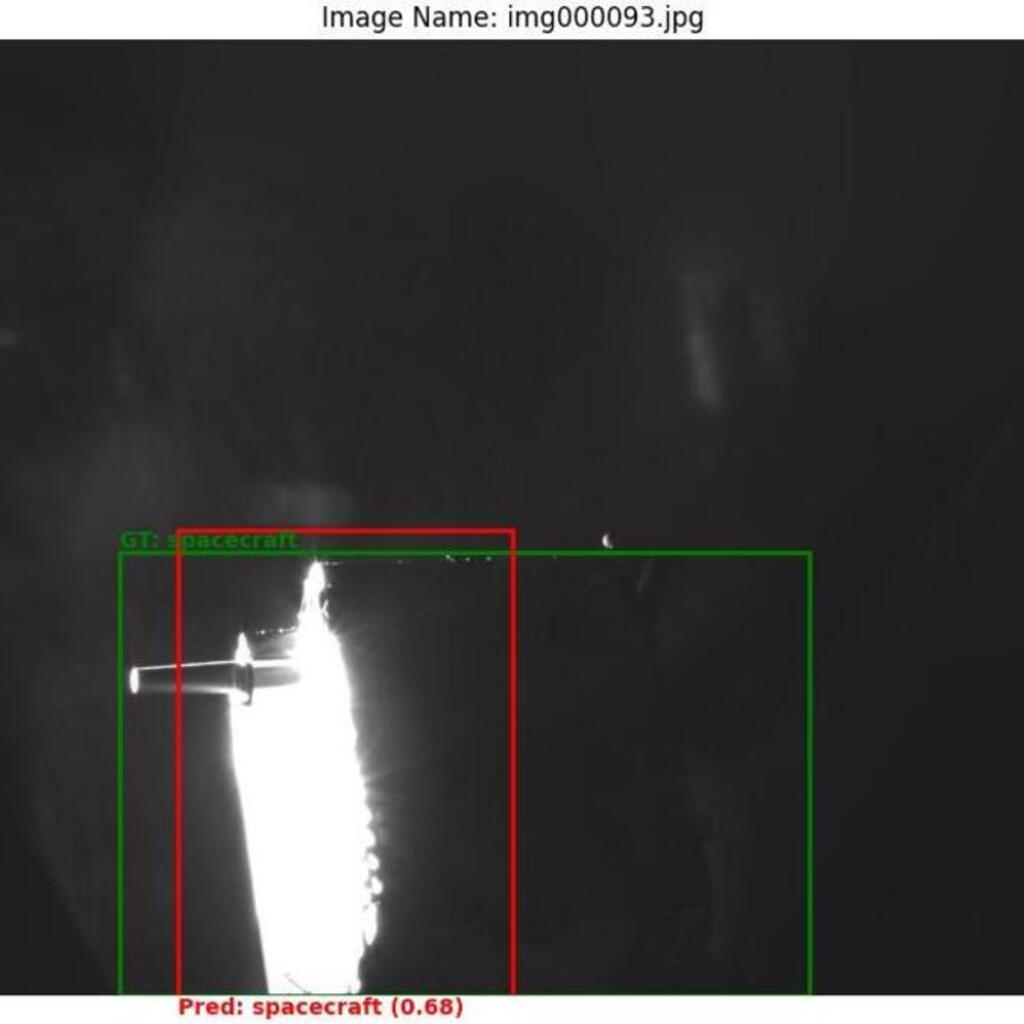} &
\includegraphics[width=\myw,  ,valign=m, keepaspectratio,] {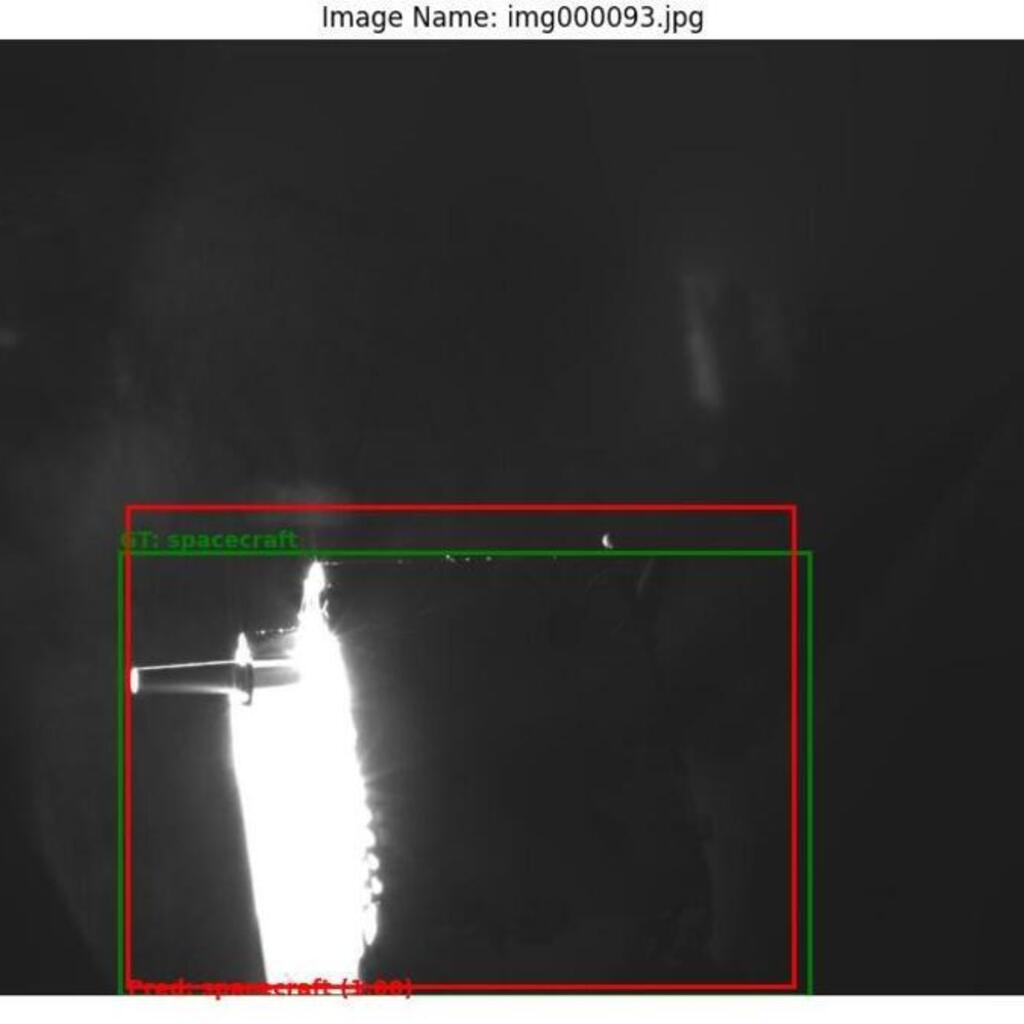} &

\includegraphics[width=\myw,  ,valign=m, keepaspectratio,] {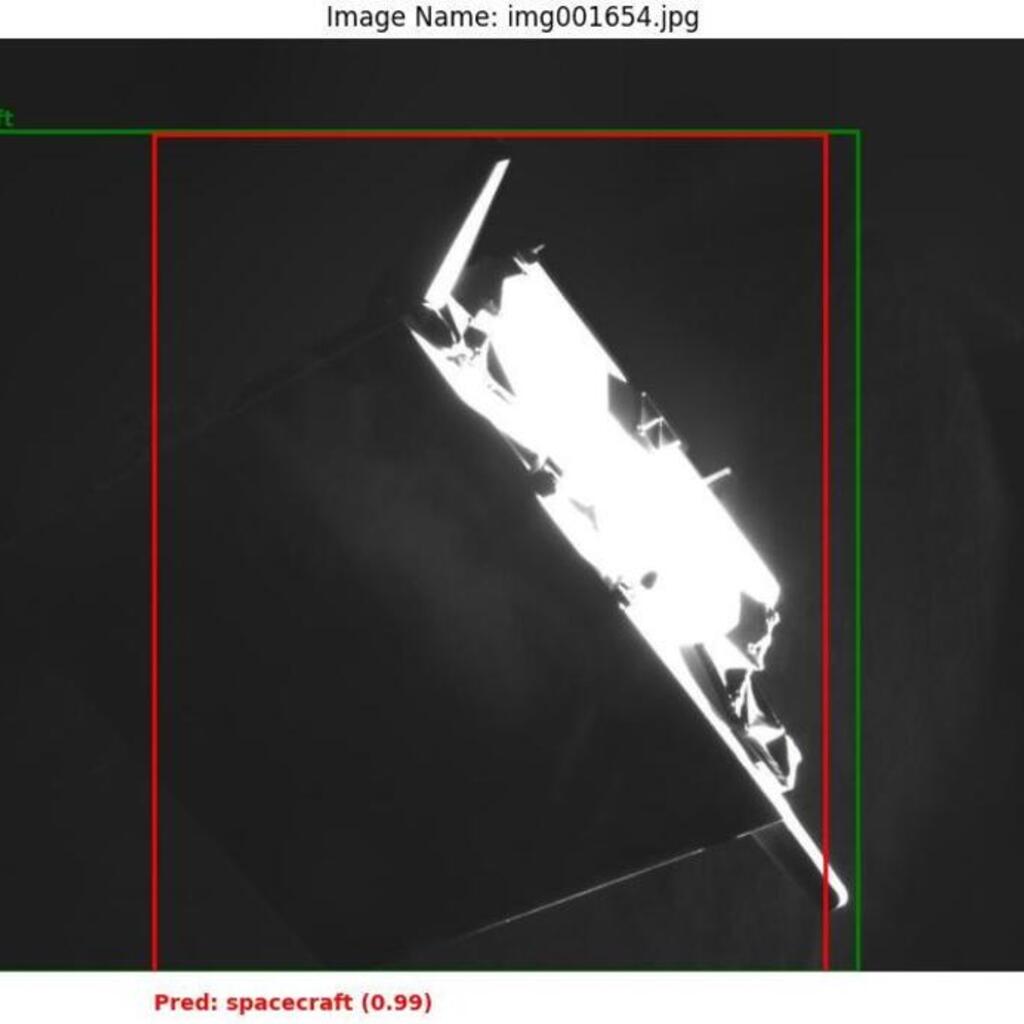} & 
\includegraphics[width=\myw,  ,valign=m, keepaspectratio,] {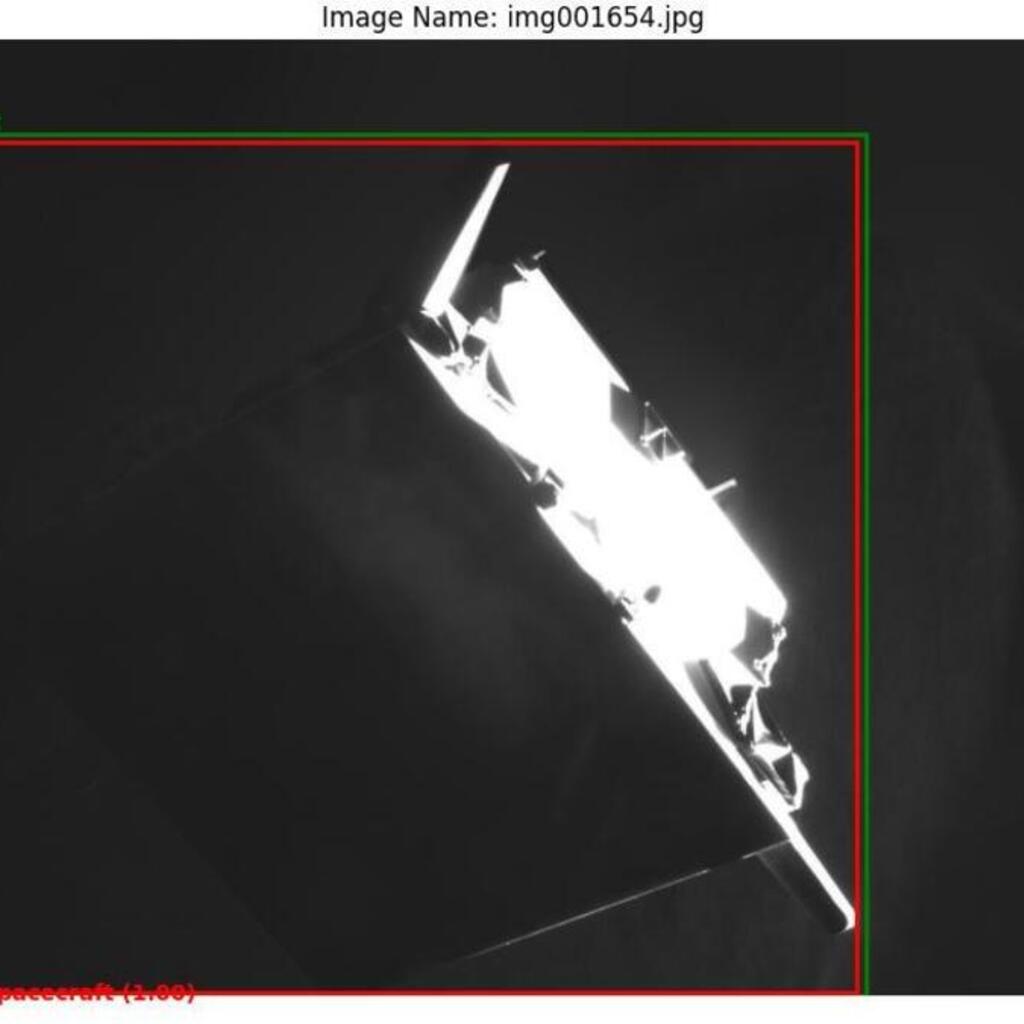}\\

{\rotatebox[origin=t]{90}{\textit{\textbf{Sunlamp}}}}  &
\includegraphics[width=\myw,  ,valign=m, keepaspectratio,] {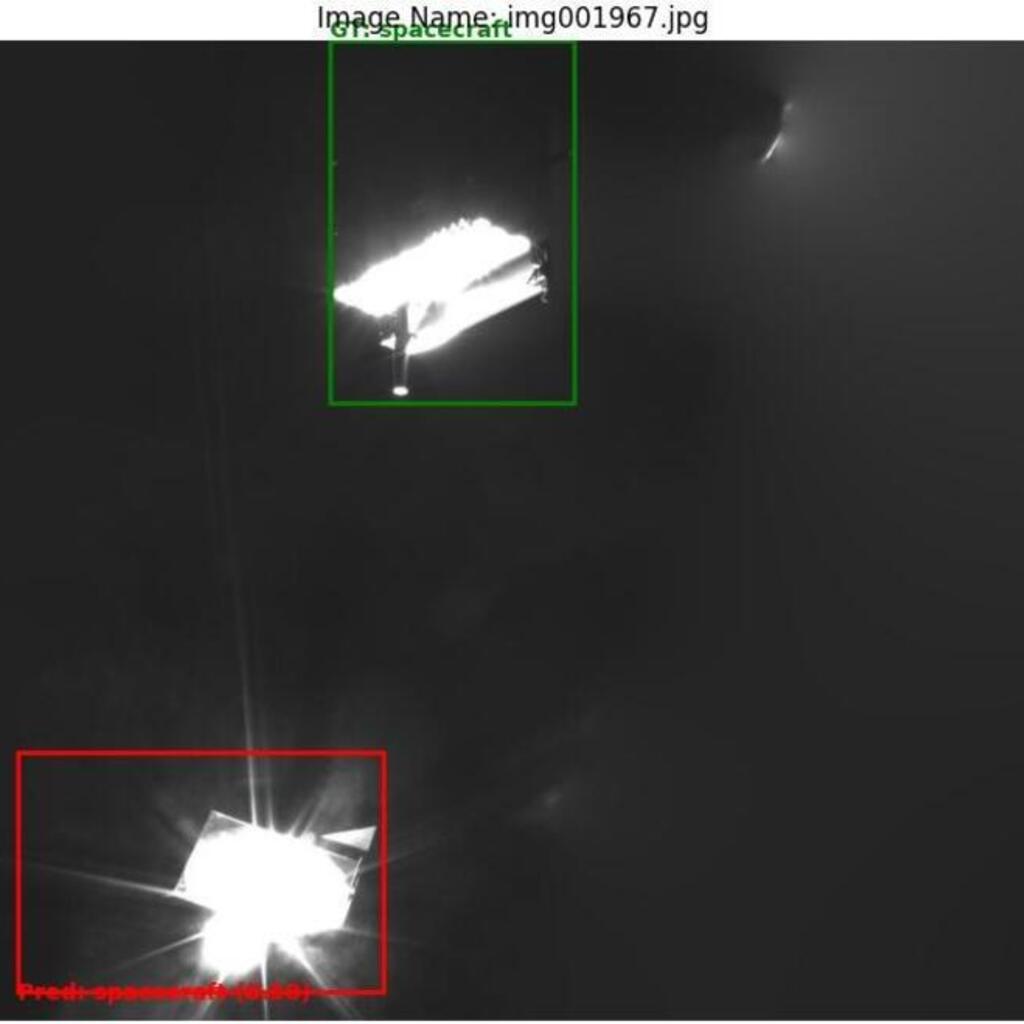} & 
\includegraphics[width=\myw,  ,valign=m, keepaspectratio,] {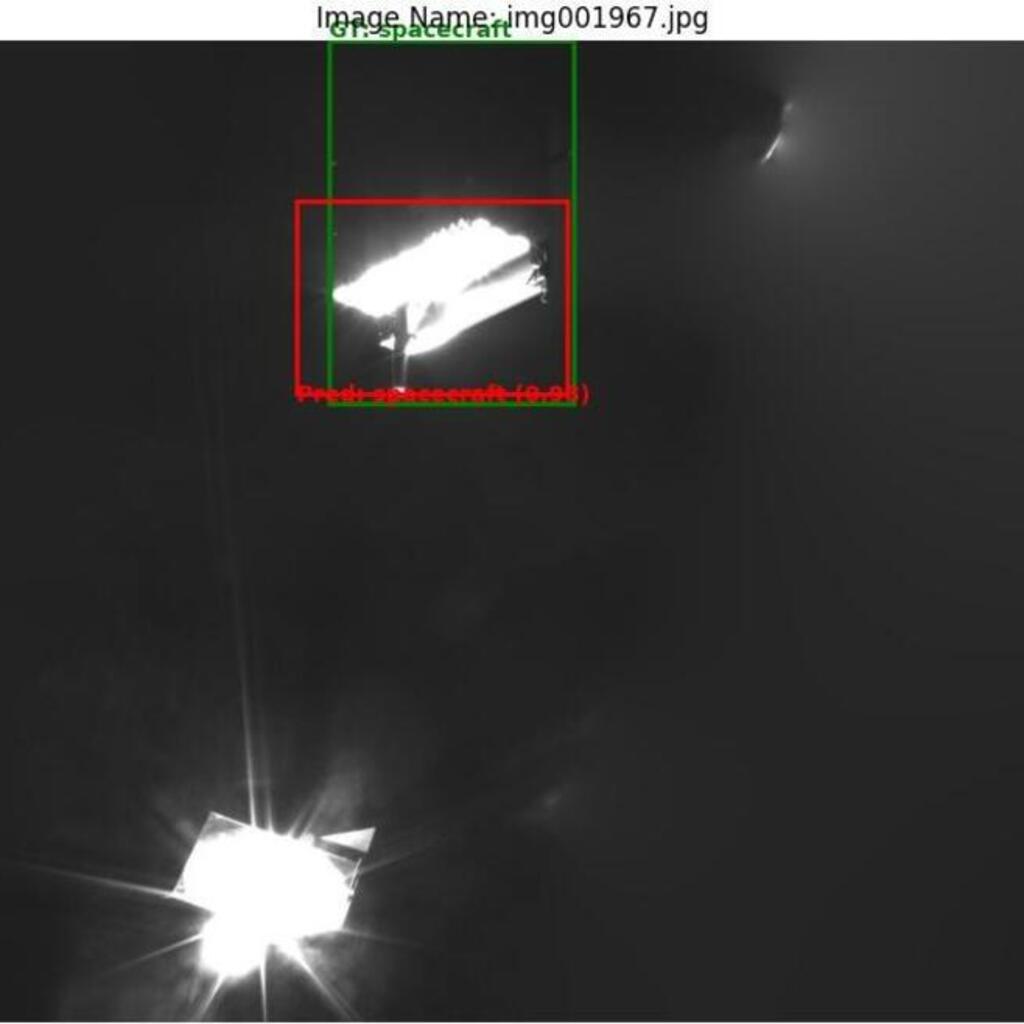} & 

\includegraphics[width=\myw,  ,valign=m, keepaspectratio,] {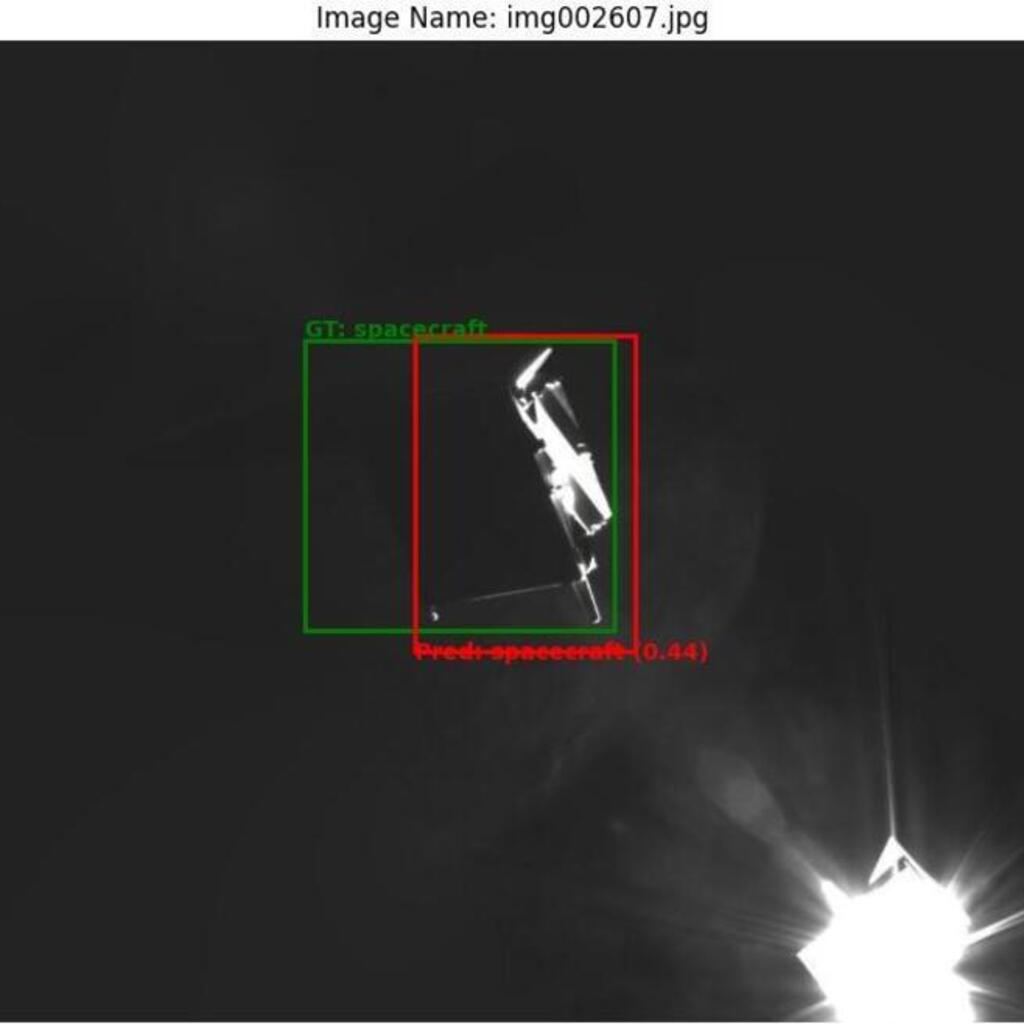} & 
\includegraphics[width=\myw,  ,valign=m, keepaspectratio,] {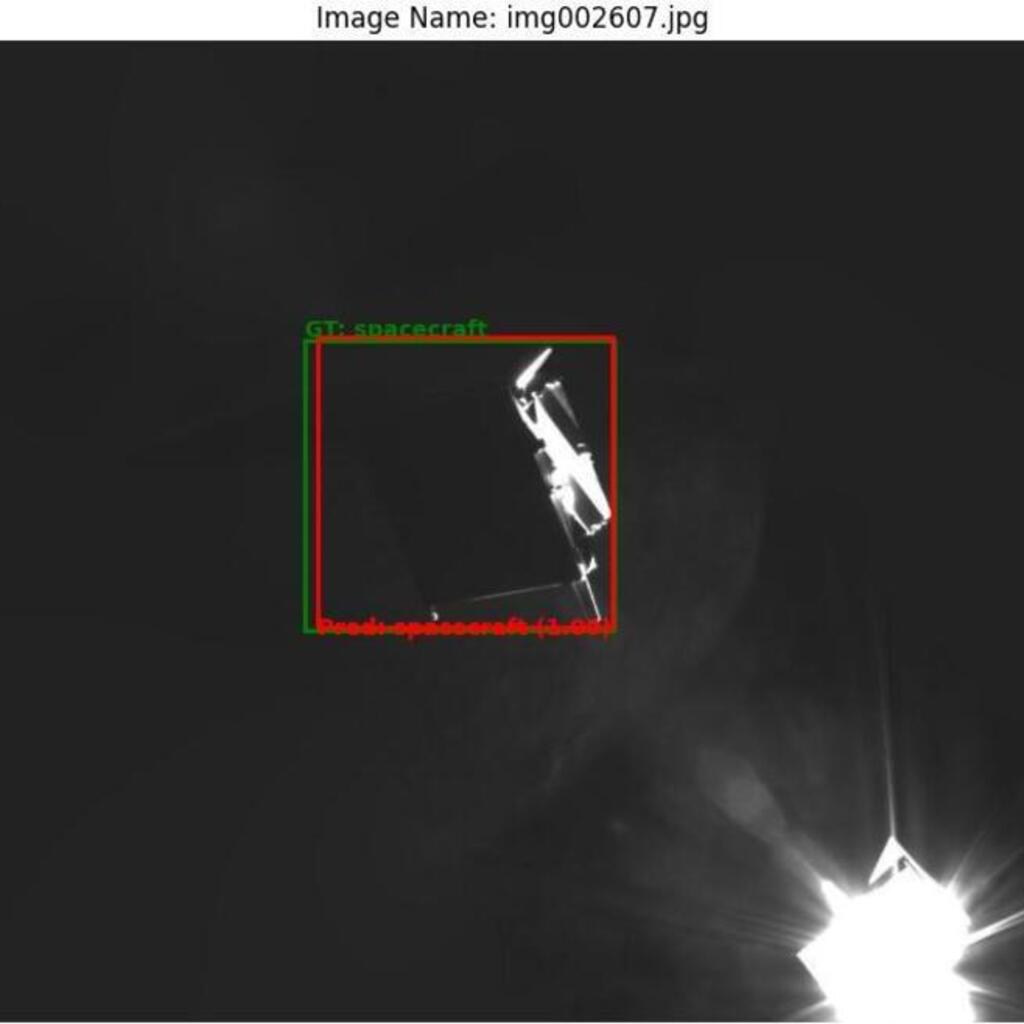}\\

{\rotatebox[origin=t]{90}{\textit{\textbf{Spark}}}}  &
\includegraphics[width=\myw,  ,valign=m, keepaspectratio,] {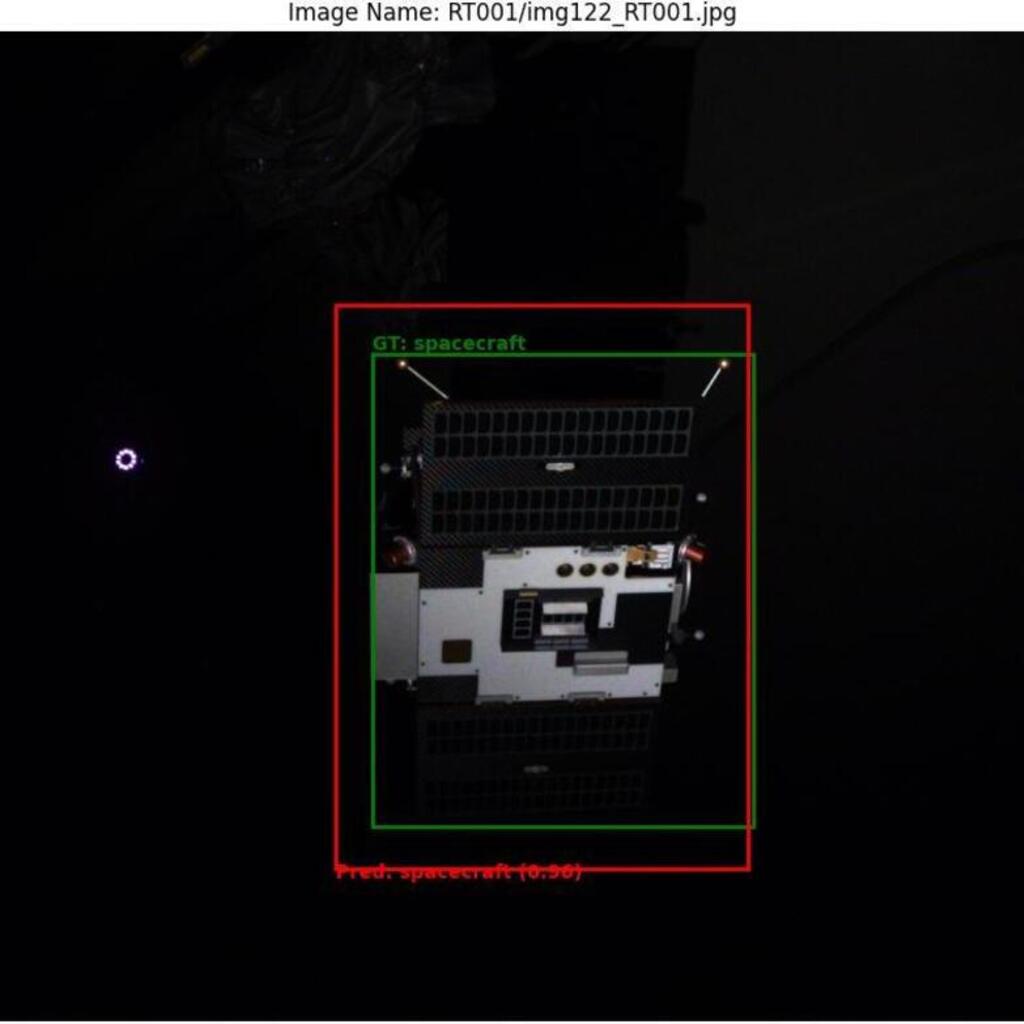} &
\includegraphics[width=\myw,  ,valign=m, keepaspectratio,] {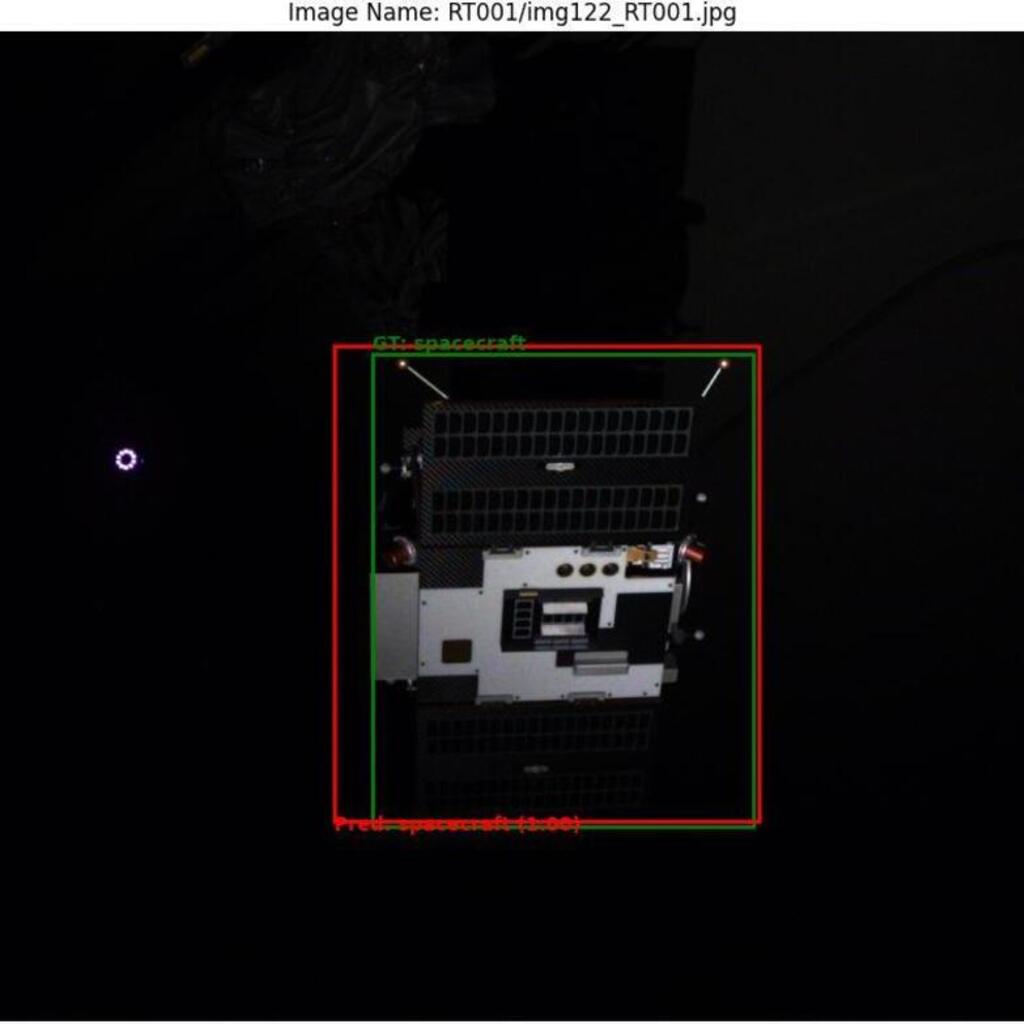} &

\includegraphics[width=\myw,  ,valign=m, keepaspectratio,] {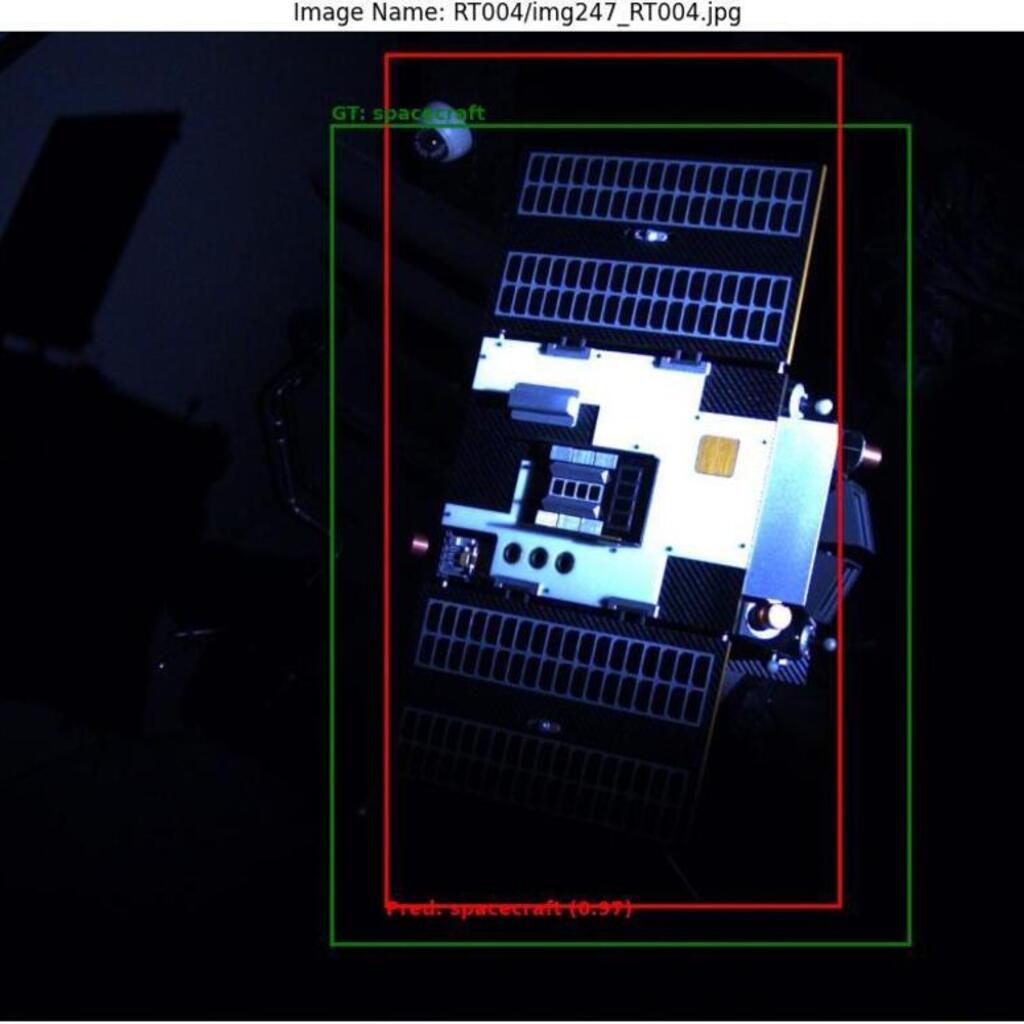} & 
\includegraphics[width=\myw,  ,valign=m, keepaspectratio,] {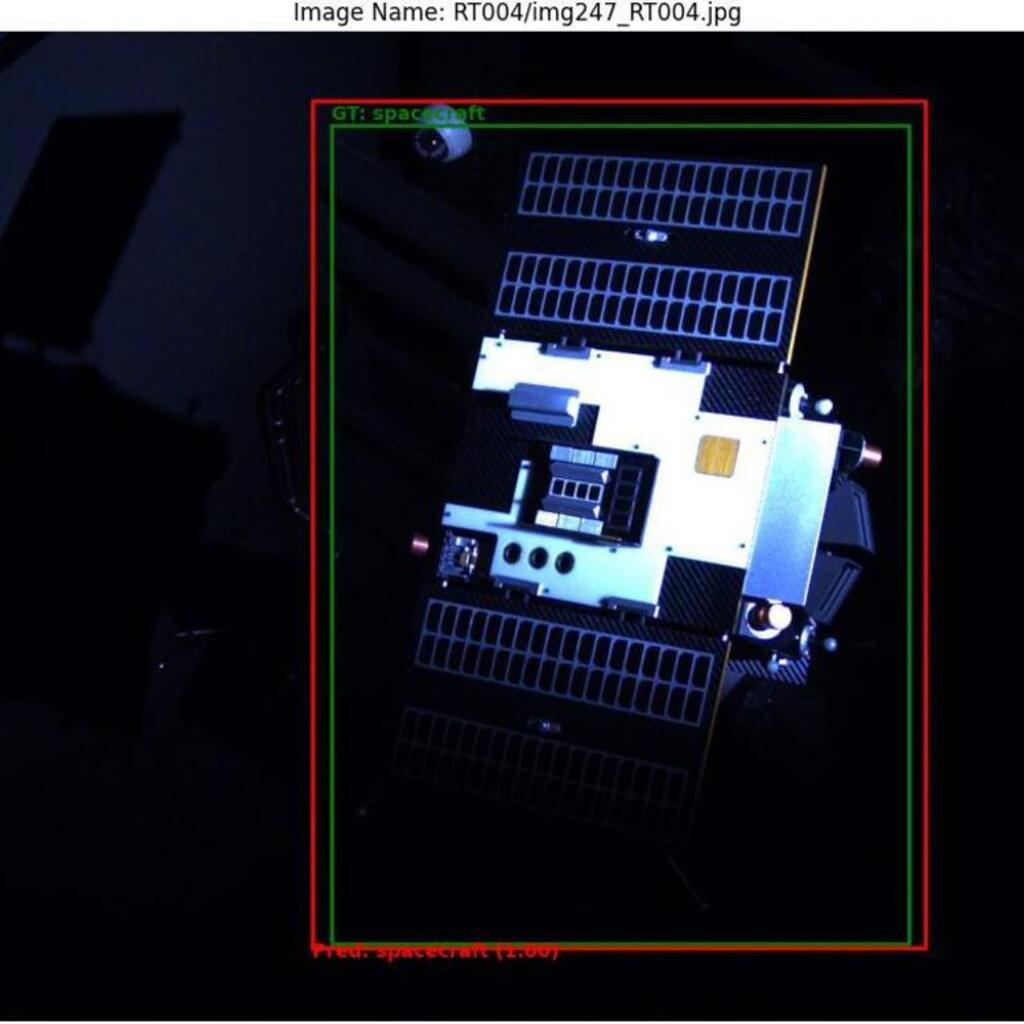}\\

{\rotatebox[origin=t]{90}{\textit{\textbf{Spark}}}}  &
\includegraphics[width=\myw,  ,valign=m, keepaspectratio,] {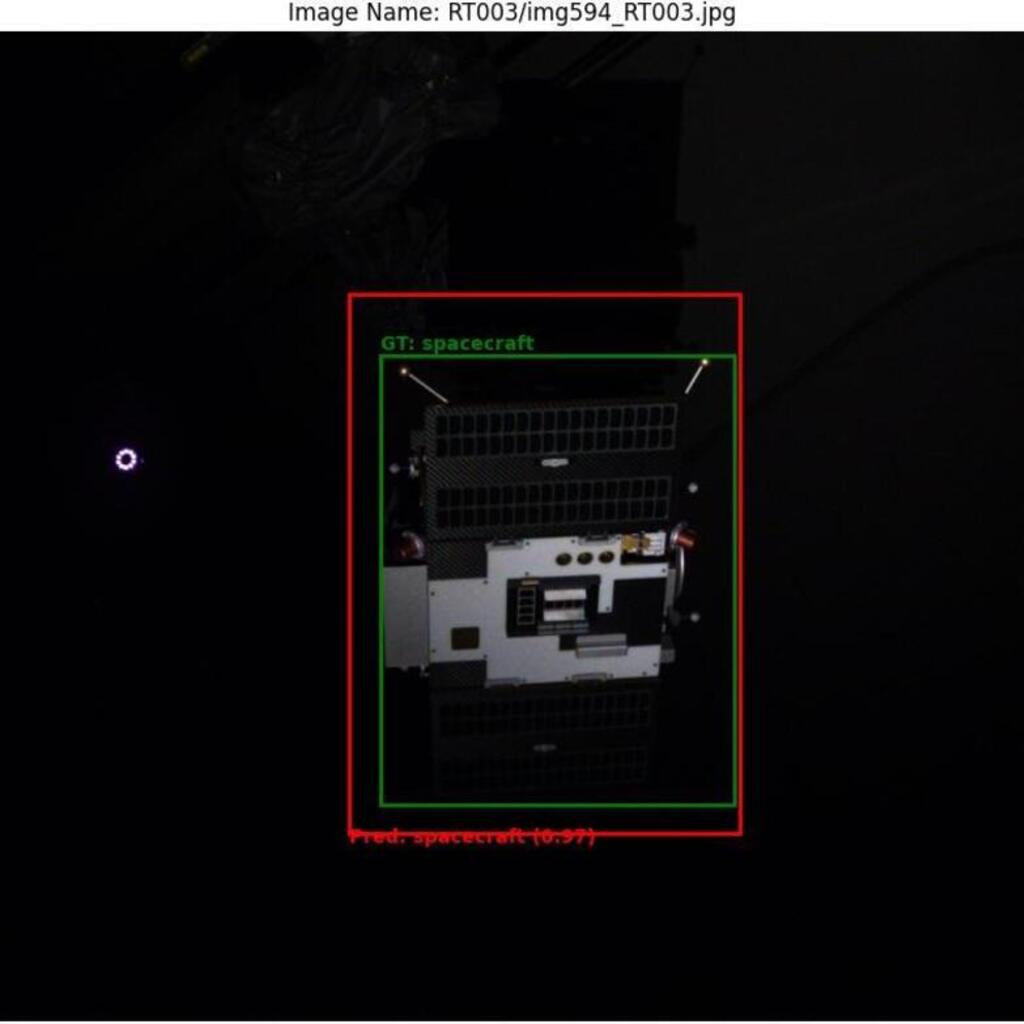} & 
\includegraphics[width=\myw,  ,valign=m, keepaspectratio,] {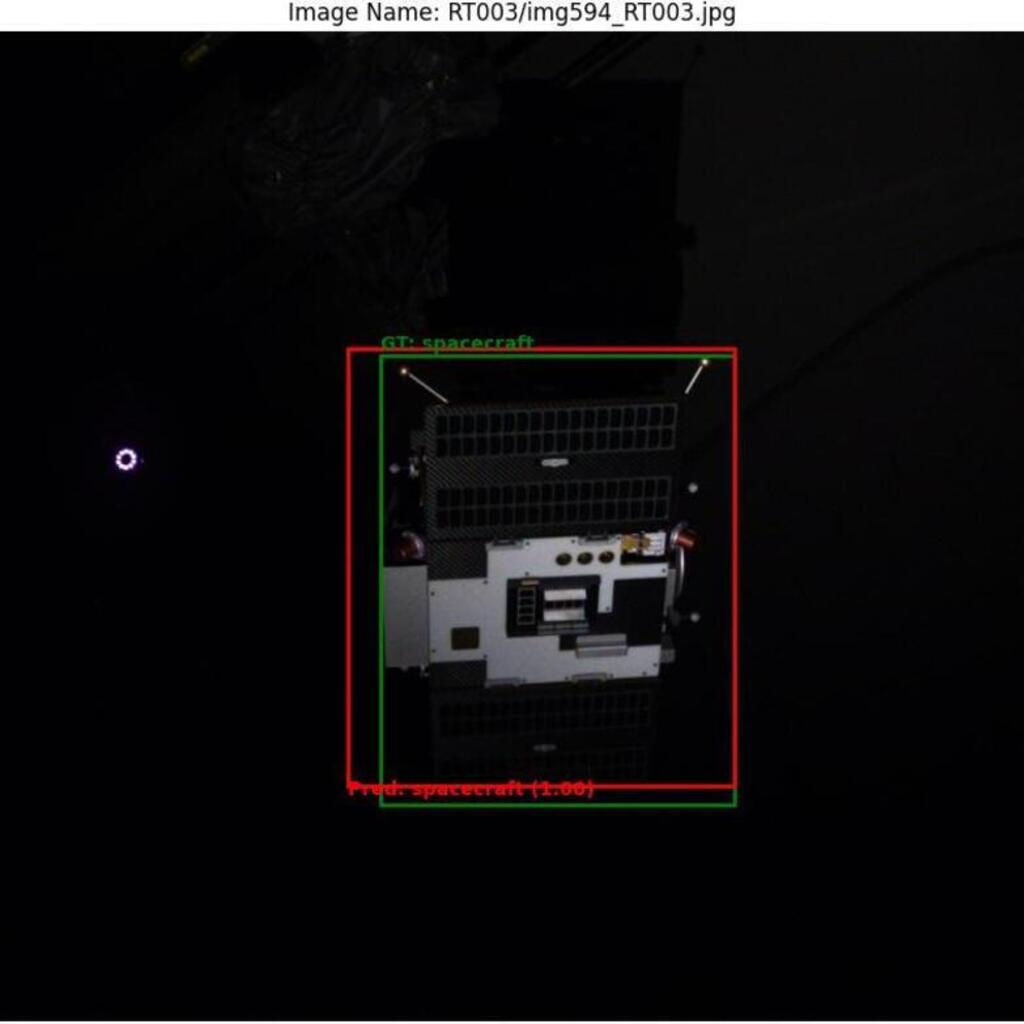} & 

\includegraphics[width=\myw,  ,valign=m, keepaspectratio,] {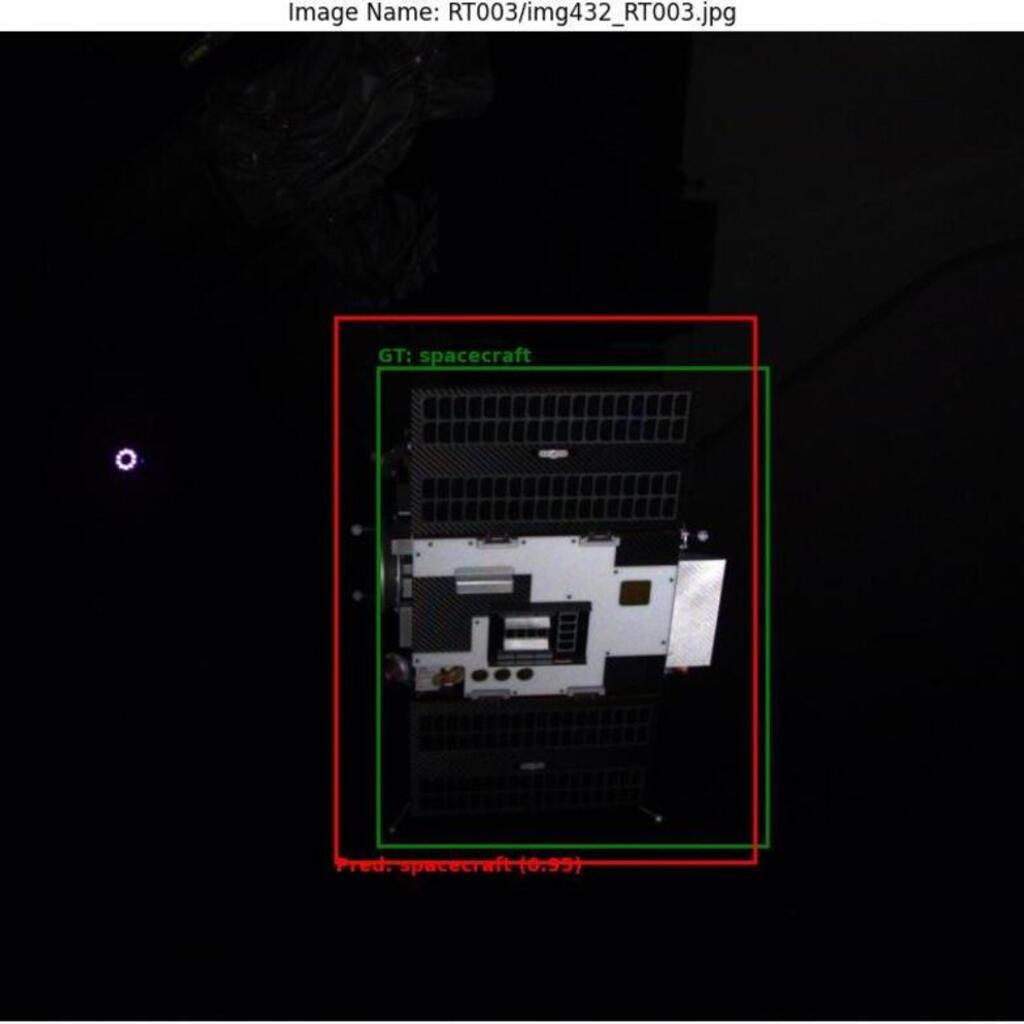} & 
\includegraphics[width=\myw,  ,valign=m, keepaspectratio,] {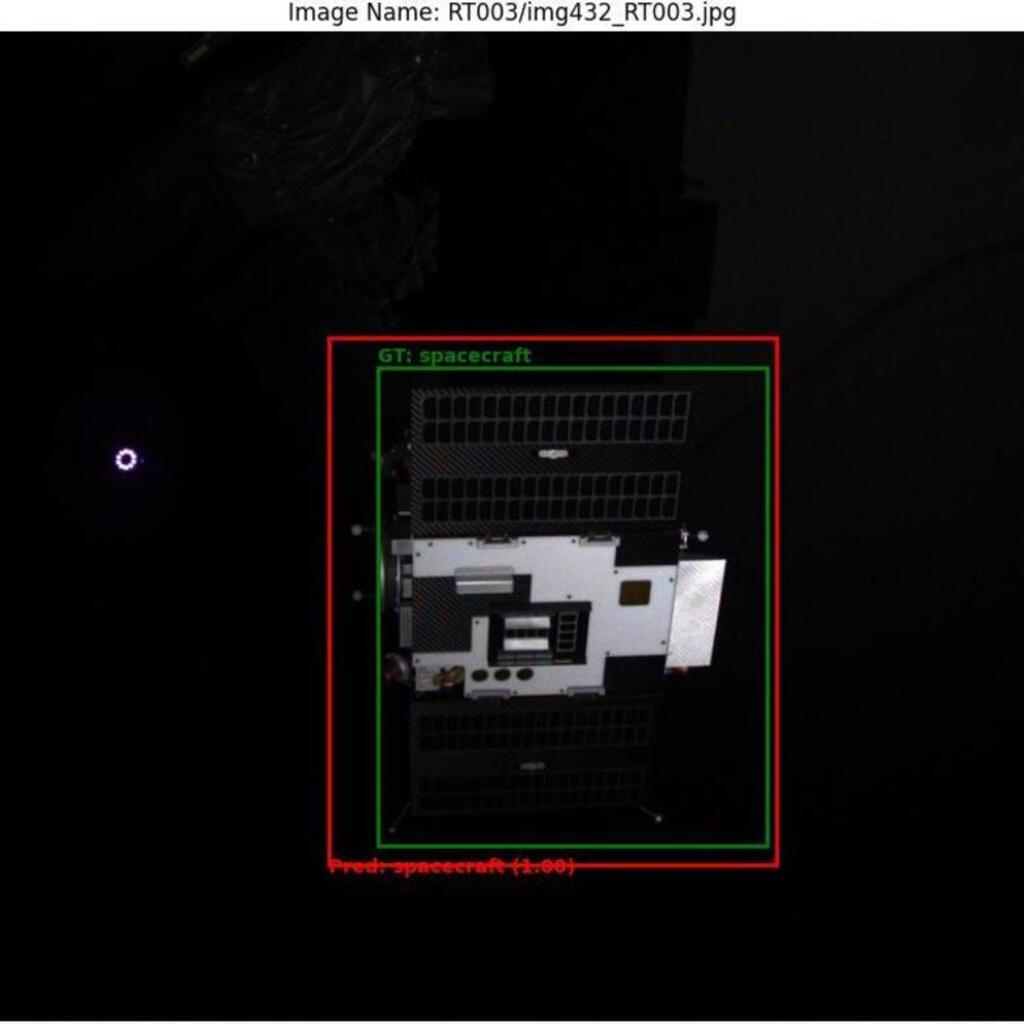}\\

\end{tabular}
\caption{Visual comparison of \textit{ORACLE} and our method on the three datasets used in experiments. Green boxes denote the ground truth and red boxes shows the predictions. Our supervised domain adaptation significantly improves the prediction performance across various cases such as earth background, another light source and close up imagery.}
\label{fig:our_results}
\end{figure*}
\renewcommand{\arraystretch}{1}

\subsection{Qualitative Results}

We present visual results of \textit{ORACLE} and our method in Figure~\ref{fig:our_results} for all datasets. Our supervised domain adaptive object detection method significantly improves the \textit{ORACLE} performance in all datasets by getting affected from the background as in the first two rows and external light sources as in the fourth row.

\section{Conclusion}
\label{sec:conc}

We presented a supervised domain adaptation approach for spacecraft object detection using limited real-world annotations. By combining domain-invariant feature learning and invariant risk minimization, our method improves performance under domain shift with minimal supervision. Experiments on two space datasets and two detector architectures show up to 20-point gains in average precision with only 200 labeled real images. These findings demonstrate the practical value of the method for real-world space applications with constrained annotation resources.

\section*{Acknowledgments}
Authors would like to thank Inder Pal SINGH, Jose SOSA, Dan PINEAU, and Nidhal Eddine CHENNI for great discussions.
This research has been conducted in the context of the \texttt{DIOSSA} project, supported by the European Space Agency (ESA) under contract no. \texttt{4000144941241 NL/KK/adu}. The experiments were performed on the Luxembourg National Supercomputer MeluXina. The authors gratefully acknowledge the LuxProvide teams for their expert support and the partners of the consortium for their contributions.

\newpage
\clearpage
{
  \small
  \bibliographystyle{ieeetr}
  \bibliography{main}
}

\end{document}